# TOWARDS MACHINE LEARNING ON DATA FROM PROFESSIONAL CYCLISTS

Agrin Hilmkil[1], Oscar Ivarsson[1], Moa Johansson[1], Dan Kuylenstierna[1], Teun van Erp[2]

[1]Chalmers University of Technology, Sweden.

[2]Vrije Universiteit Amsterdam, The Netherlands.

**Abstract**

Professional sports are developing towards increasingly scientific training methods with increasing amounts of data being collected from laboratory tests, training sessions and competitions. In cycling, it is standard to equip bicycles with small computers recording data from sensors such as power-meters, in addition to heart-rate, speed, altitude etc. Recently, machine learning techniques have provided huge success in a wide variety of areas where large amounts of data ("big data") is available. In this paper, we perform a pilot experiment on machine learning to model physical response in elite cyclists. As a first experiment, we show that it is possible to train a LSTM machine learning algorithm to predict the heart-rate response of a cyclist during a training session. This work is a promising first step towards developing more elaborate models based on big data and machine learning to capture performance aspects of athletes.

## Introduction

Professional sports have long been developing towards increasingly scientific training methods. One of the most fundamental principles in sport science and training practices is the dose-response relationship, i.e., the physical adaption to training stimulus. At first sight, it seems simple; more training better performance; but in reality, there are several stimuli and the responses are not linear. It is a complex multi-dimensional nonlinear problem. An early system-level description of the challenge was reported in (Calvert et. al. 1976). Beside the challenge of multi-dimensional nonlinear modelling, another issue is to define accurate and valid input data.

Recently development in the ICT sector has supported athletes and coaches with vast amount of data being available from GPS sensors, heart rate monitors, power meters, and motion sensors. Collected over time it becomes a lot of data that can be used to design training programs and follow up training loads with target to peak performance with reduced risk of overreaching, which otherwise may result in injuries, sickness, and degraded performance. Still, among practitioners rather primitive methods are used to handle the large data sets. Coaches and athletes mostly rely on gut feeling or studies of single parameters, e.g., heart rate or power data. At best, the levels are normalized to maximum capacity or critical power. Coggan and Allen have defined the training stress score (TSS) (Allen & Coggan, 2010), which is a key number that is calculated from real-time work intensity related to each subject documented critical intensity, defined through either power, heart rate or speed. Despite being widely used for assessment of training, TSS still considers only one variable, e.g., the power. The nonlinear intensity relationships in the TSS formula are also ad-hoc without any clear motivation about the nonlinear relationship used (Kuylenstierna, 2018). Recently, TSS compared to TRIMP have also been shown to differ significantly depending whether the data was collected in a training or racing session (van Erp, 2018).

One of the challenges in modelling physiological and biological systems is the large number of stimulus involved making it impossible to clearly define the governing dynamics needed for an analytical model starting from first principles. Under these constraints a black box modelling approach is more adequate (Cooper 1991).

With the rise of computing power and availability of large amounts of digital data, *machine learning* has revolutionised many fields, from machine vision (Cireşan et al, 2012) to natural language translation (Sutskever et al. 2014) and genetics (Chicco et al, 2014), just to mention a few. There exists a wide range of machine learning algorithms. Here, we are mainly concerned with supervised learning algorithms, which typically have in common that they model complex statistical relationships, implemented as adjustable weights in the model. These weights are typically gradually adjusted to fit a set of *training data*, containing labelled examples (for instance pictures of animals labelled by their species), followed by an evaluation on unseen validation data to assess how well the model generalises to novel examples.

There has been some work on machine learning methods in the context of performance analysis in sports. In (Pfeiffer & Hohmann, 2011), neural network algorithms are used to several rather different tasks: predicting talent development in young swimmers, recognising tactical patterns in handball matches and finally to analyse the effect of training on performance in one Olympic swimmer. They found that neural network-based approaches performed better than alternative models. In (Churchill, 2014) the aim was to predict fitness indicators of elite cyclists based on data from training and races, collected from bike non-laboratory setting such as bike computers and self-assessments in training diaries. The motivation was to aid athletes and coaches approximate current fitness values without having to conduct time consuming laboratory tests, which is often infeasible during racing season due to time constraints. A challenge here was that the data was relatively sparse, only coming from a few cyclists. Small datasets are a problem in machine learning. A novel algorithm specially developed to handle small datasets, the so-called hybrid artificial neural network ensemble model (HANNEM), was proposed and used for modelling field data from a few world class cyclists.

In our work, we consider data in the form of time-series collected from bike computers and associated sensors recorded during training sessions of professional cyclists. We thus had access to both a larger volume and longer time-series of data than e.g. (Churchill, 2014). Furthermore, we apply modern recent so called *deep-learning* methods, which have revolutionised many applications of machine learning. To our knowledge this is the first study on deep-learning methods applied to sports data. Specifically, we use a so called *long-short term memory (LSTM) neural network model* (Hochreiter & Schmidhuber, 1997), which is suitable for the analysis of time series data. Our pilot study is an experiment to train an LSTM model to predict the cyclists heart-rate at any given point in time. While this is not our ultimate goal from a performance analysis point of view, it is a suitable evaluator for assessing if machine learning at all can *capture and encode* the dependencies between different physiological and performance factors.

**Methods**

For this study we had access to a dataset collected from both male and female professional cyclists. The files in the dataset each correspond to one training session or race and was extracted from the cyclist's bike computers. These small devices record one data point per second, including information from various sensors, e.g. heart rate monitor, power meter (if the bike is equipped with one), GPS position, altitude, speed etc. As a typical training session or race lasts between 2-6 hours, each file consists of time series of thousands or tens of thousands of data-points.

The quality of the data varied between different individuals, with some always riding with a heart rate monitor and power meter, while others rarely rode with sensors on the bike. Furthermore, the sensors are not always measuring perfectly and some sessions contained obviously spurious values. The dataset also contained more historical data for the male cyclists. Due to these variations, we ended up only using data from 15 male cyclists for the experiments described here.

In addition to the data from bike computers, we also had access to some meta-data about each cyclist. This included for instance some information about performance in laboratory test such as maximum oxygen uptake and power output and heart rate at lactate threshold etc.

As a first experiment in applying machine learning to performance data from cyclists we decided to attempt to train a machine learning algorithm to predict the heart rate at any given point in time during a training session. We first give a brief overview of the kind of machine learning algorithms we used, followed by a more detailed description of how we trained and evaluated a model for predicting heart rate.

*A Brief Introduction to Recurrent Neural Networks and LSTMs*

Each training session in our dataset is represented as a time series, as described above. To train a machine learning algorithm on such data, we need a suitable algorithm which can remember and utilise earlier information to solve some task at the present point. *Long short-term memory* networks (LSTMs for short) (Hochreiter & Schmidhuber, 1997), is a machine learning model in the family of *recurrent neural networks* (RNNs for short) (Lipton et al. 2015) which has been successfully applied to a variety of tasks involving data in the form of sequences or time-series, including speech recognition, translation (Sutskever et al. 2014) and image captioning. Like all recurrent neural network models, LSTMs consist of a chain of repeated processing units, taking an input each, and feeding the result of its computation to the next unit in the network. What is special about LSTMs is that they also keep track of a *state* (effectively its "memory"), which is updated by each unit and then read by the next unit in the network. This way, LSTMs can "remember" information from much earlier in the input sequence (here earlier points in time, as we are handling time series), which other types of RNNs will not do well.

Training a neural network, such as an LSTM, essentially consists of repeatedly calculating a so-called *loss function* for each data point in the training set, and then adjusting and fine-tuning the *weights* connecting the units in the network as to minimise the error between the current and expected values. The weights essentially regulate what information is passed through the network, how much importance the network should attach to it, and what is forgotten or ignored. For example, in our experiment to predict the heart rate at a given point in time, the loss function is the difference between the actual heart rate, and the heart rate computed by our network. For each training step, this error is propagated through the network and weights are updated in order to try to make this error smaller. The process is repeated many times for a large number of training examples. Typically, there are a number of parameters for training that are tweaked experimentally, and many different algorithms for training different network architectures. See (Lipton et al. 2015) for an overview.

The process of training a neural network model such as an LSTM can be computationally expensive and time consuming, but the situation has much improved in recent years with the help of specialised software libraries, for example TensorFlow (https://www.tensorflow.org), and specialised hardware called Graphical Processing Units (GPUs), which thanks to parallelism perform well for training tasks.

When using machine learning algorithms, the available data is typically divided into a larger training set and a smaller evaluation set. The training set is used to train the chosen model, often under a longer period of time, and possibly over several passes over the data, allowing the model to adjust and learn through tweaking its internal parameters. To ensure the model has learned parameters that generalise also to new data, the performance of the model is evaluated on the unseen evaluation set.

*Predicting Heart Rate with an LSTM*

Given the successes found in the literature, and the nature of our data as a time series we decided to attempt to implement a LSTM network and train it to predict heart rate of a cyclist. While this is not our ultimate goal, it serves as a good first indicator to evaluate if the machine learning model manages to learn some representation of physical response. A trained LSTM network cannot only be used for prediction, but also for producing compact encodings of the data, as numeric vectors called *embeddings* (Srivatava et al, 2015). Such embeddings can be useful for instance for visualisation of the data. In Figure 1, we show the intended workflow of our system.

The source code which was used to run these experiments is available online (see https://github.com/agrinh/procyclist_performance) and is based on the TensorFlow library written in Python. The target of this experiment was to predict heart rate for any time step, given input variables:

- Time (seconds)
- Speed (km/h)
- Distance (km)
- Power (Watt)
- Cadence (pedal strokes/minute)
- Power/weight (Watt/kg)
- Altitude (m)
- Heart rate 30 s. prior to current time (beats per minute)

The raw data contained one sample per second. With long training sessions lasting up to 5-6 hours, each session could be extremely long. For practical reasons, the data was therefore down-sampled with a factor of 1/0.3.

As a pre-processing step for cleaning the raw data, we removed sessions where no heart rate values were recorded, or where spurious values were recorded (e.g. heart rates way over 200, indicating faulty readings of the HR-monitor). Also, sessions with negative values for distance, speed or power (indicating faulty readings) were removed together with sessions where the power was recorded but the distance did not increase (e.g. sessions on stationary bikes). The data was also trimmed to remove leading and trailing zeros in heart rate, indicating the rider removing the heart rate monitor with the bike computer still recording. The input parameters listed above were standardised (rescaled) to mean 0 and standard deviation 1 to avoid introducing artificial bias in any variables, as units vary between the different inputs (this is standard in machine learning applications).

The final dataset used for this experiment thus consisted of 7541 sessions from 15 male professional cyclists. Finally, as is common practice in machine learning, the data was split into a training set, consisting of data from 10 cyclists (5179 sessions), and a validation set, consisting of data from 5 cyclists (2362 sessions). After training, the model was evaluated on the unseen validation set.

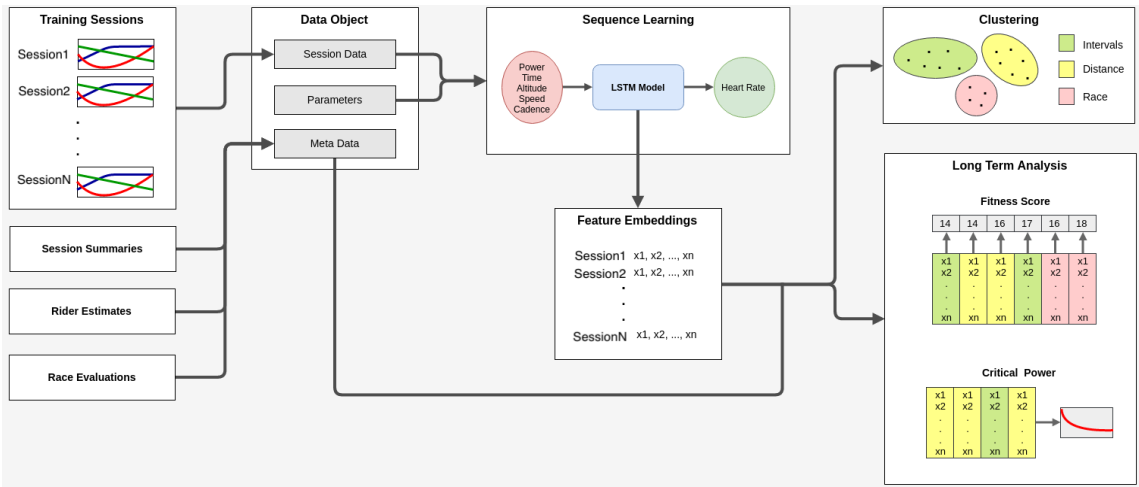

*Figure 1: Overview of the workflow of our machine learning experiment. Data from training sessions are fed to a LSTM model which is trained to predict heart-rate based on modelling physical response for the given input parameters. In addition, we can extract so called feature embeddings from the model (numeric vectors), which is a compact encoding which can be computed for a training session fed to the model. In the future, we would like to investigate in more detail if these feature embeddings form clusters with any sensible meaning. We hypothesise that similar training sessions (e.g. interval sessions vs distance sessions) will have similar embeddings. Such information could then be useful for long-term analysis.*

**Results**

The LSTM-network was trained for approximately 1 week on a Titan X Pascal GPU and obtained a final min, mean and max Root Mean Squared Error (RMSE) of 2.51, 5,62 and 25.67 on the validation set. Figure 2 and Figure 3 show our trained models predictions on two samples from the validation set, where the RMSE was less than 4. The model's predictions are very close to the actual heart rate measured both for an interval training session, where the heart rate varies quite drastically, and on a session ridden at high but rather constant velocity and power. This indicates that the LSTM-network has indeed managed to learn a representation for the cyclists' physical response. In Figure 4, we show a session where the model's prediction has a larger RMSE. Towards the end of the session there are a few short steep spikes, followed by a drop, which our model does not capture well, possibly due to the fact that it has not learnt to react to very short sprints or similar. It is also likely that the heart rate monitor has recorded some spurious values (lost contact) towards the end when it drops to near zero.

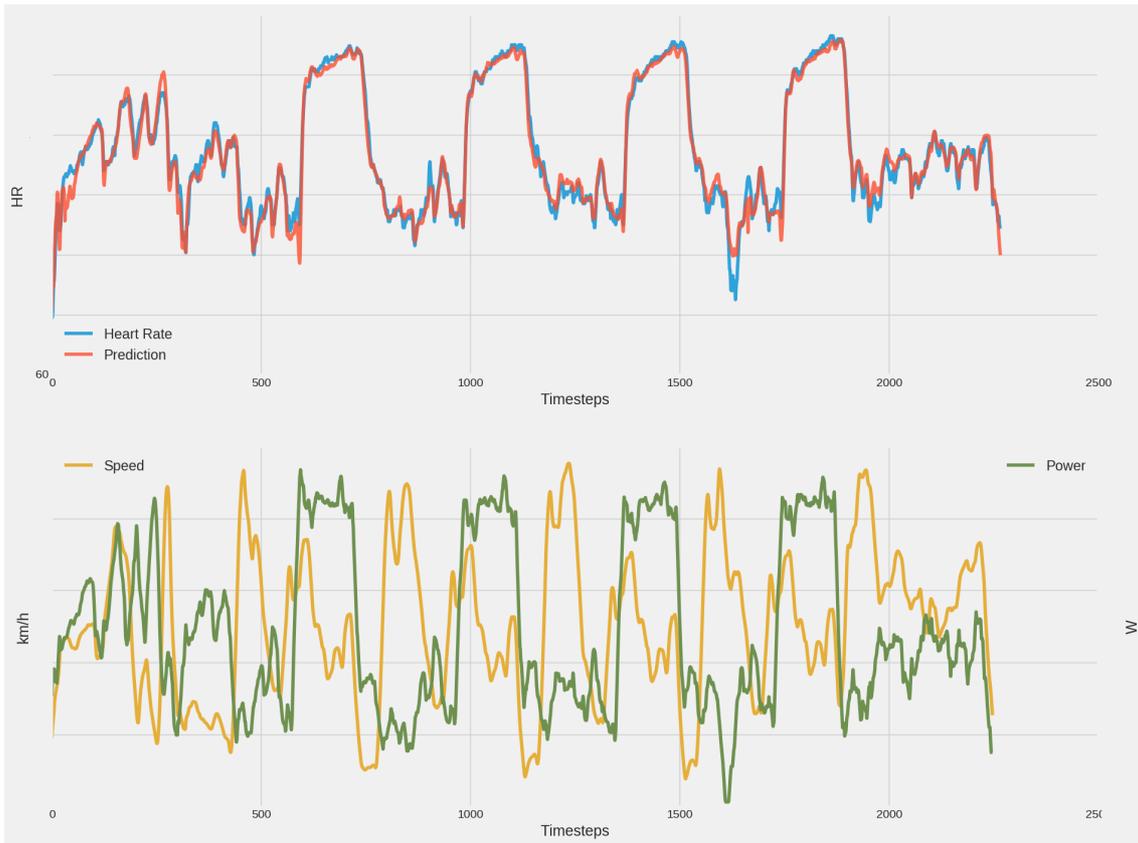

Figure 2: Prediction of heart-rate on an interval training session.

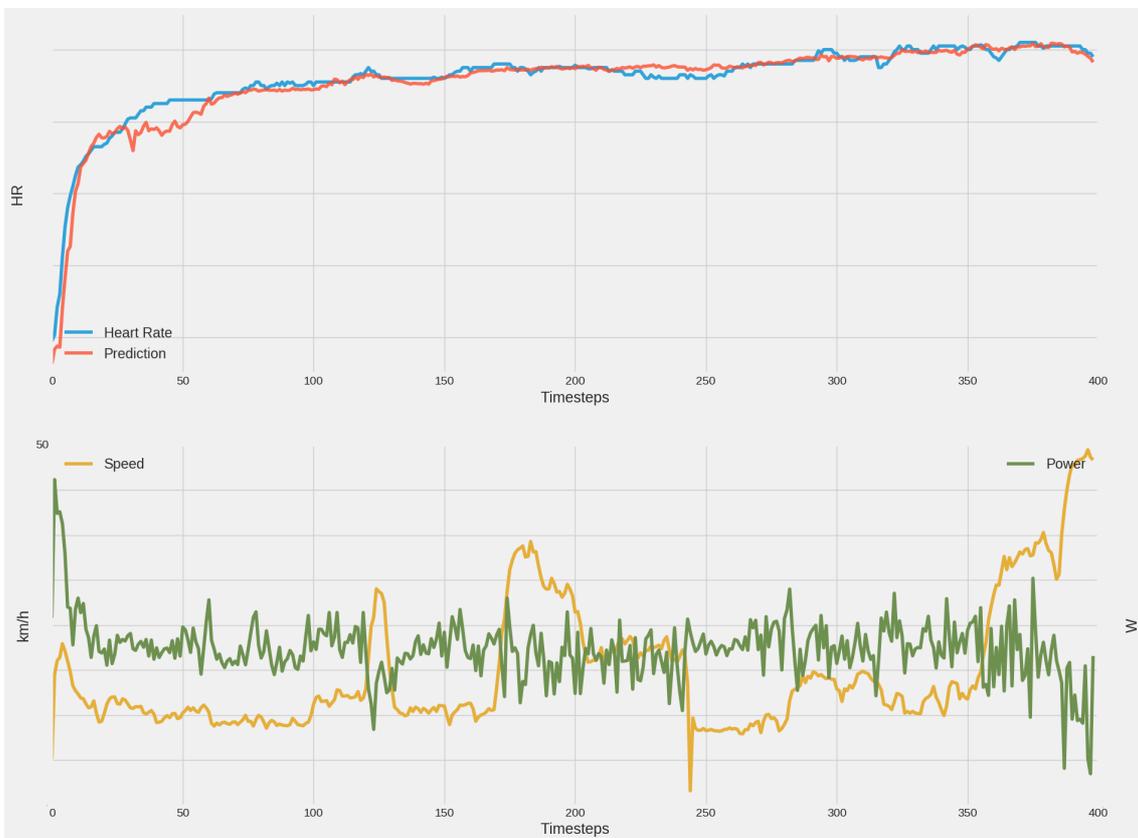

Figure 3: Prediction of heart-rate on a session at relatively even pace.

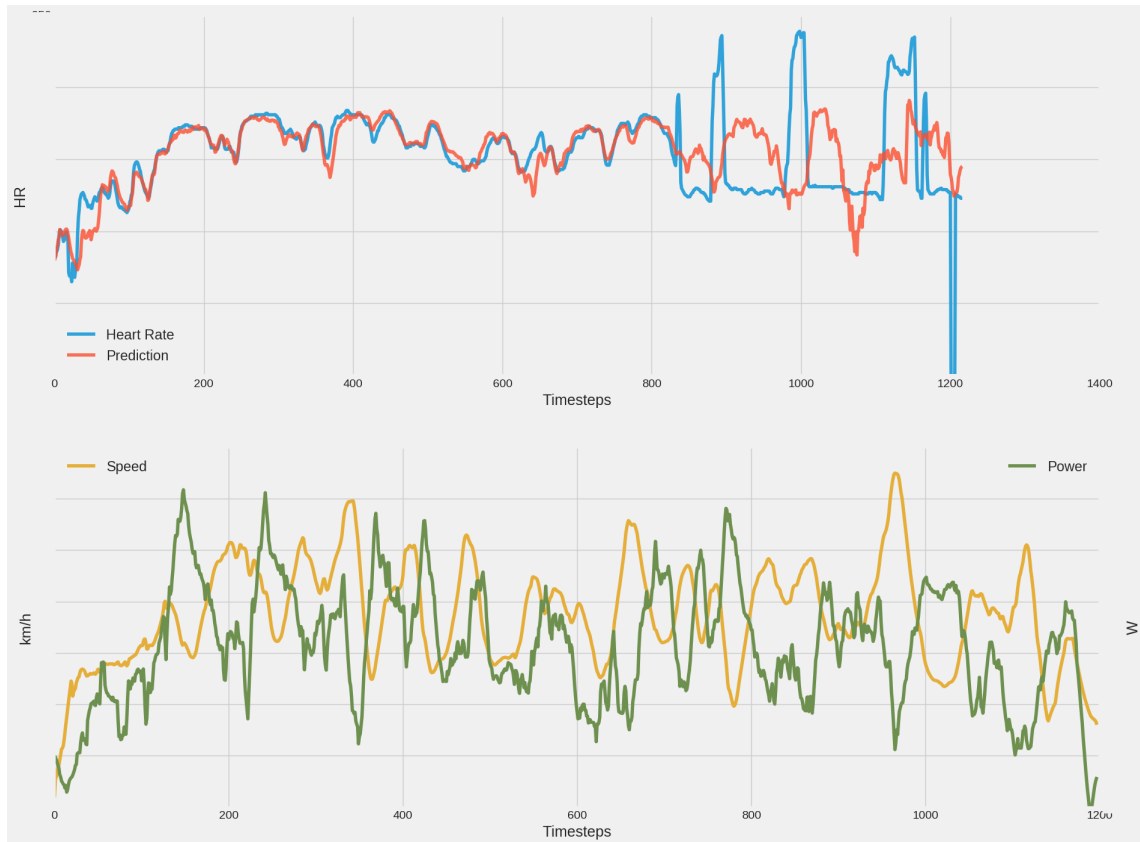

*Figure 4: A session where our model's RMSE > 4, and the predicted values differ.*

Recall that the trained LSTM model also can be used to extract compact encodings of its inputs as numeric vectors, which in turn can be used to visualise the data. The idea is that input data given to the model with similar features gets encoded to numeric vectors which are "closer" together in a multi-dimensional space, forming clusters, as is shown in Figure 5. We speculate that the embeddings possibly could be used to cluster and classify similar training sessions (interval session of varying intensity, races, time-trials, distance rides etc.). As the figure shows, there are indeed clusters being formed, but a detailed analysis of the sessions in the respective clusters is further work.

**Discussion**

We note that the time series representing training sessions are very long (several hours) compared to for instance the time series used when training machine learning models to recognise objects in short video clips (seconds). Even after down-sampling our data, each time series is still very long. This meant that training our model was somewhat time consuming, preventing us to explore all modifications to the learning parameters we would have liked within the project's time frame. For instance, experiments with systematically excluding different input parameters and observing the re-trained model's performance on the evaluation set would give us an indication of which of the inputs are most relevant for prediction. We would also like to swap input/output parameters and explore how this affects the model, e.g. predicating power given heart rate instead. This is further work.

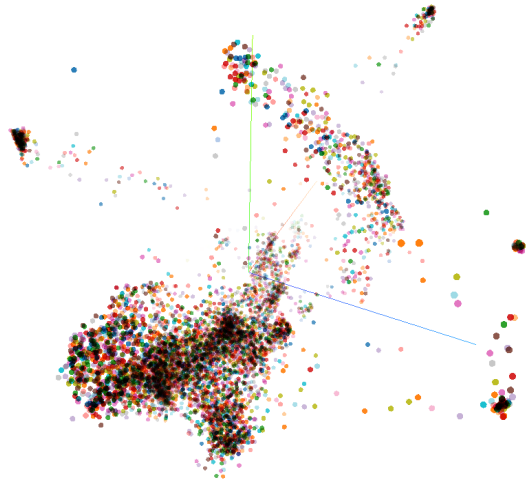

*Figure 5: Feature embeddings extracted from our LSTM model. Each plot represents a training session.*

Finally, a difference compared to conventional applications of deep learning is that while our data was long, the training set was not particularly big compared to for instance image recognition systems which typically are trained on sets containing at least tens of thousands of training examples. In future work, we will consider techniques to compensate for the relatively few but long training samples. For instance, one approach might have been to split each training session into smaller chunks, thus getting more but shorter data. However, one would then of course loose information about length of the session, which indeed can be relevant for performance metrics.

**Conclusion**

This paper describes a first experiment in applying advanced machine learning methods on performance data collected from professional cyclists, using an LSTM-model. We showed that is indeed possible to train such a model to predict the heart rate of a cyclist at any given point in time. Our work on machine learning and performance data is still at an early stage. While the heart rate prediction model we use as a first case study in this paper might not be of immediate use in performance analytics, it still indicates that further investigation into similar models are worth pursuing. As a next step, we would like to train a model to recognise for instance some measure of intensity (e.g. TSS or normalised power), possibly also including data from rider's subjective experience as recorded in their training diaries. Unlike heart rate, which is mainly a local property, these measures need to capture a representation of the whole training session. The weights of the trained machine learning model would then be forced to learn a compact encoding of the whole time-series from a session. We speculate that the model then could be used to for instance automatically classify sessions and compute training intensity measures both over an individual session, as well as over a longer training period. We would also like to experiment with other machine learning architectures, not just LSTMs.

There is interest in conducting similar experiments with data from other sports. For instance, the Chalmers University spin-off company Skisens AB (http://skisens.com) are developing power meters for ski poles and are interested in analysing data collected from cross-country skiers. In summary, applying machine learning to performance data is a new promising area of research, which has the potential to provide coaches and athletes with novel software tools to help analyse, categorise and steer training. Our work is a first step in this direction, suggesting several interesting topics to explore next.